%% file: ms.tex
\documentclass[10pt,twocolumn,letterpaper]{article}

\usepackage{cvpr}
\usepackage{times}
\usepackage{epsfig}
\usepackage{graphicx}
\usepackage{amsmath}
\usepackage{amssymb}
\usepackage{soul,color}
\usepackage{amsmath}
\usepackage[ruled,lined]{algorithm2e}
\usepackage{tabularx}
\usepackage{booktabs}
\usepackage{epstopdf}
\usepackage{color, colortbl}
\usepackage{makecell}
\usepackage{multirow}

\usepackage[pagebackref=true,breaklinks=true,letterpaper=true,colorlinks,bookmarks=false]{hyperref}

\cvprfinalcopy 

\def\cvprPaperID{9726} 

\ifcvprfinal\fi
\begin{document}

\title{Learn2Perturb: an End-to-end Feature Perturbation Learning to Improve Adversarial Robustness}

\author{Ahmadreza Jeddi\textsuperscript{\rm 1},
        Mohammad Javad Shafiee\textsuperscript{\rm 1}\\
        Michelle Karg\textsuperscript{\rm 2}, 
        Christian Scharfenberger\textsuperscript{\rm 2},
        Alexander Wong\textsuperscript{\rm 1} \\
\textsuperscript{\rm 1}Waterloo AI Institute, University of Waterloo, Waterloo, Ontario, Canada\\
\textsuperscript{\rm 2}ADC Automotive Distance Control Systems GmbH, Continental, Germany\\
\textsuperscript{\rm 1}\{a2jeddi, mjshafiee, a28wong\}@uwaterloo.ca \\ 
\textsuperscript{\rm 2}\{michelle.karg, christian.scharfenberger\}@continental-corporation.com 
}

\maketitle
\thispagestyle{plain}
\pagestyle{plain}

\begin{abstract}
  While deep neural networks have been achieving state-of-the-art performance across a wide variety of applications, their vulnerability to adversarial attacks limits their widespread deployment for safety-critical applications.  Alongside other adversarial defense approaches being investigated, there has been a very recent interest in improving adversarial robustness in deep neural networks through the introduction of perturbations during the training process. However, such methods leverage fixed, pre-defined perturbations and require significant hyper-parameter tuning that makes them very difficult to leverage in a general fashion. In this study, we introduce Learn2Perturb, an end-to-end feature perturbation learning approach for improving the adversarial robustness of deep neural networks.  More specifically, we introduce novel perturbation-injection modules that are incorporated at each layer to perturb the feature space and increase uncertainty in the network.  This feature perturbation is performed at both the training and the inference stages.  Furthermore, inspired by the Expectation-Maximization, an alternating back-propagation training algorithm is introduced to train the network and noise parameters consecutively. Experimental results on CIFAR-10 and CIFAR-100 datasets show that the proposed Learn2Perturb method can result in deep neural networks which are 4-7\% more robust on $l_{\infty}$ FGSM and PDG adversarial attacks and significantly outperforms the state-of-the-art against $l_2$ C\&W attack and a wide range of well-known black-box attacks.
\end{abstract}

\section{Introduction}
The vulnerability of DNN models to adversarial examples have raised major concerns~\cite{akhtar2018threat,carlini2018audio,cheng2018seq2sick,moosavi2016deepfool,sharif2016accessorize} on their large-scale adaption in a wide variety of applications. 

Adversarial attacks can be divided into two categories of black-box and white-box attacks based on the level of information available to the attacker. Black-box attacks usually perform queries on the model, and they have partial information regarding the data and the structure of the targeted model~\cite{ilyas2018black,papernot2017practical}. On the other hand, white-box attacks have a better understanding of the model that they attack to; therefore, they are more powerful than black-box attacks~\cite{he2019parametric,szegedy2013intriguing}. This understanding might vary between different white-box attack algorithms; nonetheless, gradients of the model's loss function with respect to the input data is the most common information utilized to modify input samples and generate adversarial examples. First-order white-box adversaries are the most common attacking algorithms which only use the first order of gradients \cite{carlini2017towards,madry2017towards,moosavi2016deepfool,szegedy2013intriguing} to craft the adversarial perturbation.

In the realm of defense mechanisms, approaches like distillation \cite{papernot2016effectiveness,papernot2016distillation}, feature denoising \cite{xie2019feature}, and adversarial training \cite{goodfellow2014explaining,madry2017towards} have been proposed to resolve the vulnerability of DNNs on adversarial attacks. Adversarial training is considered as a very intuitive yet very promising solution to improve the robustness of DNN models against adversarial attacks.

Madry {\it et al.}~\cite{madry2017towards} illustrated that adversarial learning using Projected Gradient Descent (PGD) for generating on-the-fly adversarial samples during training can lead to trained models which provide robustness guarantees against all first-order adversaries. They  experimentally showed  that the adversarial examples in a $l_{\infty}$ ball distance around the original sample with many random starts in the ball generated with PGD, all have approximately the same loss value when are fed to the network as input. Due to this fact, they provide the guarantee that as long as the attack algorithm is a first-order adversary, the local maximas of the loss value would not be significantly better than those found by PGD.

Applying regularization techniques is another approach to train more robust network models~\cite{bietti2019kernel}. To do so, either new loss functions were proposed with added or embedded regularization terms (i.e., adversarial generalization)~\cite{elsayed2018large,hein2017formal,sun2016depth} or the network was augmented with new modules~\cite{he2019parametric,lecuyer2018certified,liu2018towards,liu2018adv} for regularization purposes making the network more robust at the end. 

Randomization approaches and specifically random noise injection~\cite{he2019parametric,liu2018towards,liu2018adv} has  been recently proposed as one of the network augmentation methods to address the adversarial robustness in deep neural networks. A random noise generator as an extra module is embedded in the network architecture adding random noise to the input or the output of layers. Although the noise distribution usually follows a Gaussian distribution for its simplicity, it is possible to use different noise distributions. This noise augmentation technique adds more uncertainty in the network and makes the adversarial attack optimization harder which improves the robustness of the model.

While the noise injection technique has shown promising results, determining the parameters of the distribution and how to add the noise values to the network are still challenging. The majority of the methods proposed in literature~\cite{lecuyer2018certified,liu2018towards,liu2018adv, zhang2019theoretically} manually select the parameters of the distribution. However, He {\it et al.}~\cite{he2019parametric} recently proposed a new algorithm in which the noise distributions are learned in the training step. Their proposed Parametric Noise Injection (PNI) technique injects trainable noise to the activations or the weights of the CNN model. The problem associated to the proposed PNI technique is that the noise parameters tend to converge to zero as the training progresses, making the noise injection progressively less effective over time. This problem is partially compensated through the utilization of PGD adversarial training as suggested by Madry {\it et al.}~\cite{madry2017towards}, but the decreasing trend of noise parameter magnitudes still remains, and thus, limits the overall effect of the PNI.

In this paper, the Learn2Perturb framework, an end-to-end feature perturbation learning approach is proposed to improve the robustness of DNN models. An alternating back-propagation strategy is introduced where the following two steps are performed in an alternating manner: i) the network parameters are updated in the presence of feature perturbation injection to improve adversarial robustness, and ii) the parameters of the perturbation injection modules are updated to strengthen perturbation capabilities against the improved network. Decoupling these two steps helps both sets of parameters (i.e., network parameters and perturbation injection modules) to be trained to their full functionalities and produces a more robust network. To this end,  our contributions can be folded as below:
\begin{itemize}
    \item A highly efficient and stable end-to-end learning mechanism is introduced to learn the perturbation-injection modules to improve the model robustness against adversarial attacks. The proposed  alternating back-propagation method inspired by Expectation-Maximization (EM) concept trains the network and noise parameters in a consecutive way gradually without any significant parameter-tuning effort.
    
    \item A new effective regularizer is introduced to help the network learning process which smoothly improves the noise distributions. Combining this regularizer and PGD-adversarial training helps the proposed Learn2Perturb algorithm achieve the state-of-the-art performances.
    
    \item Exhaustive experiments are conducted for various white-box and black-box adversarial attacks on CIFAR-10 and CIFAR-100 datasets, and new state-of-the-art performances are reported for these algorithms.

\end{itemize}

The paper is organized as follows: section~\ref{LR} provides a discussion of related work in terms of  different adversarial attacks; the proposed Learn2Perturb approach is presented in Section~\ref{sec:method} and experimental results and discussion are presented in Section~\ref{sec:res} followed by a conclusion.

\section{Related Work}
\label{LR}
The gradients of the loss function with respect to the input data are very common information used by adversarial attack algorithms. In this type of approaches,  the proposed algorithms try  to maximize the loss value of the network by crafting the minimum perturbations into input data. 

Fast Gradient Sign Method (FGSM) \cite{szegedy2013intriguing} is the simplest yet a very efficient white-box attack. For a DNN parametrized with $W$ (i.e., where the network is encoded as $f_W(x)$) and loss function $\mathcal{L}$, for any input $x$, the  FGSM attack computes the adversarial example $x'$ as:
\begin{align}
    x' = x + \epsilon \cdot \text{sign}\Big(\nabla_x \mathcal{L}\big(f_W(x),x\big)\Big)
\end{align}
where $\epsilon$ determines the attack strength and $\text{sign}(\cdot)$ returns the sign tensor for a given tensor. Using this gradient ascent step, FGSM tries to locally maximize the loss function $\mathcal{L}$. 

The FGSM approach is extended by projected gradient descent (PGD)~\cite{kurakin2016adversarial_scale,madry2017towards} where for a number of $k$ iterations, PGD produces $x_{t+1} = \text{bound}_{l_{p}}(FGSM(x_{t}), x_0)$, in which $x_{0}$ is the original input and $0 \leq t \leq k-1 $. Using projection, the $\text{bound}_{l_{p}}(x', x)$ simply ensures that $x'$ is within a specified $l_{p}$ range of the original input $x$.

Madry {\it et al.}~\cite{madry2017towards} illustrated that different PGD attack restarts, each with a random initialization for input within the \mbox{$l_{\infty}$--ball} around $x$, find different local maximas with very similar loss values. Based on this finding, they claimed that PGD is a universal first-order adversary.

C\&W attack~\cite{carlini2017towards} is another strong first-order attack algorithm which finds perturbation $\delta$ added to input $x$ by solving the optimization problem formulated as:
\begin{align}
  min \Big[ ||\delta||_{p} + c \cdot f (x + \delta)\Big]  \;\;s.t.\;\;\;  x + \delta \in [0, 1]^{n}  
  \label{eq:cw}
\end{align}
where $p$ shows the norm distance. While $p$ can be any arbitrary number, C\&W is most effective when $p=2$; as such, here, we only consider $l_{2}$-norm for C\&W evaluations. Moreover, $f(\cdot)$ encodes the objective function driving the perturbed sample to be misclassified (ideally $f(\cdot) \leq 0$), and $c$ is a constant balancing the two terms involved in \ref{eq:cw}. It is worth noting that, all the white-box attacks explained here (i.e. FGSM, PGD, and C\&W) are first-order adversaries.

Black-box attacks can only access a model via queries; sending inputs and receiving corresponding outputs to estimate the inner working of the network. To fool a network, the well-known black-box attacks either use surrogate networks~\cite{ilyas2018black,papernot2017practical} or estimate the  gradients~\cite{chen2017zoo,su2019one} via multiple queries to the targeted network.

In the surrogate network approach, a new network mimicking the behavior of the target model~\cite{papernot2017practical} is trained. Attackers perform queries on the target model and generate a synthetic dataset with the query inputs and associated outputs. Having this dataset, a surrogate network is trained. Recent works~\cite{liu2016delving,papernot2017practical} showed that adversarial examples fooling the surrogate model can also fool the target model with a high success rate. A simple variant of the surrogate model attack, Transferability adversarial attacks~\cite{papernot2016transferability}, is when the surrogate model has access to the same training data as the interested network.  Adversarial examples fooling the substituted network usually transfer to (fool) the target model as well. 
Since substitute networks may not always be successful~\cite{chen2017zoo,ilyas2018black}, black-box gradient estimation attacks only deal with the target model itself. Zeroth order optimization (ZOO)~\cite{chen2017zoo} and attacks alternating only a few pixels~\cite{su2019one} approaches are examples of this kind of black-box attack, to name a few.

\vspace{-0.25cm}
\section{Methodology}
\label{sec:method}
\begin{figure*}
    \centering
    \includegraphics[width=18cm]{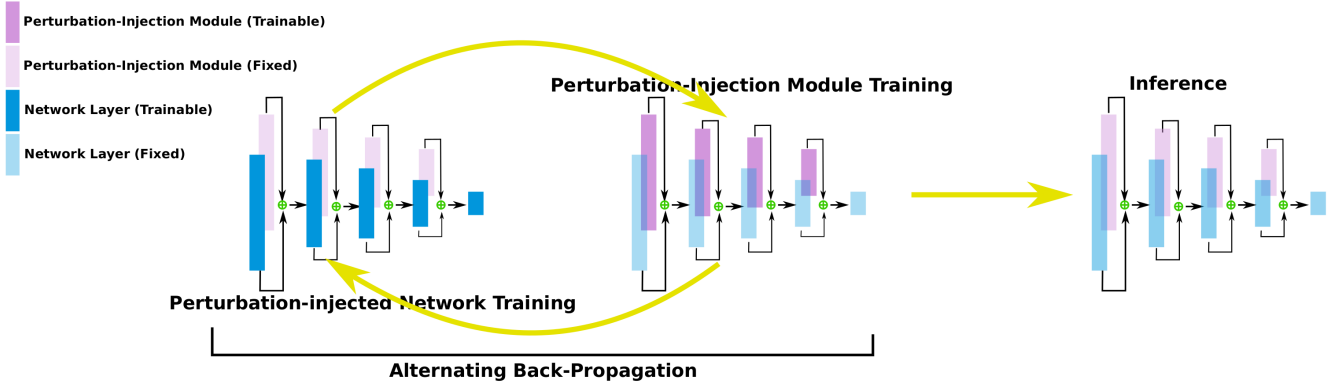}
    \caption{Overview of Learn2Perturb: During training, an alternating back-propagation strategy is introduced where the following two steps are performed in an alternating manner: i) the network parameters are updated in the presence of feature perturbation injection to improve adversarial robustness, and ii) the parameters of the perturbation injection modules are updated to strengthen perturbation capabilities against improved network.  The learned perturbation injection modules can be added to some or all tensors in the network to inject perturbations in feature space for two-prong adversarial robustness: i) improve robustness during training when training under perturbation injection, and ii) increase network uncertainty through interference-time perturbation injection to make it difficult to learn an adversarial attack.}
    \label{fig:learn2perturb}
    \vspace{-0.5cm}
\end{figure*}

In this work, we propose a new framework called  Learn2Perturb for improving the adversarial robustness of a deep neural network through end-to-end feature perturbation learning. Although it has been illustrated both theoretically and practically~\cite{araujo2019robust,pinot2019theoretical} that randomization techniques can improve the robustness of deep neural networks\footnote{Theoretical background on the effect of randomization algorithm to improve the robustness of a deep neural network model is discussed in the supplementary material.}, there is still not an effective way to select the distribution of the noise in the neural networks. In Learn2Perturb, trainable perturbation-injection modules are integrated into a deep neural network with the goal of injecting customized perturbations into the feature space at different parts of the network to increase the uncertainty of its inner workings within an optimal manner.  We formulate the joint problem of learning the model parameters and the perturbation distributions of the perturbation-injection modules in an end-to-end learning framework via an alternating back-propagation approach~\cite{han2017alternating}. As shown in Figure~\ref{fig:learn2perturb},  the proposed alternating back-propagation strategy for the joint learning of the network parameters and the perturbation-injection modules is inspired from the EM technique; and it comprises of two key alternating steps: i) \textbf{Perturbation-injected network training}: the network parameters are trained by gradient descent while the proposed perturbation-injection modules add layer-wise noise to the feature maps (different locations in the network). Noise injection parameters are fixed during this step. ii) \textbf{Perturbation-injection module training}: the parameters of the perturbation-injection modules are updated via gradient descent and based on the regularization term added to the network loss function, while network parameters are fixed.

The effect of using such a training strategy is that in step (i), the model minimizes the loss function of the classification problem when noise is being injected into multiple layers, and the model learns how to classify despite the injected perturbations. And in step (ii), the noise parameters are updated with a combination of network gradients and the regularization term applied to these parameters. The goal of this step is to let the network react to the noise injections via gradient descent and pose a bigger challenge to the network via a smooth increase of noise based on  the regularizer. The trained perturbation-injection modules perturb the feature layers of the model in the inference phase as well. 

\subsection{Perturbation-Injection Distribution}
Given the observable variables $X$, $W$ as the input and the set of weights in the neural network, respectively, the goal is to model the neural network as a probabilistic model such that the output of the model, $Y$, is drawn from a distribution rather than a deterministic function. A probabilistic output is more robust against adversarial perturbation. As such, $Y$ can be formulated as:
\begin{align}
    Y \sim P(X;W,\theta)
\end{align}
where $W$ and $\theta$ show the set of network and noise parameters, respectively, and $X$ is the input fed into the network. The output $Y$ is drawn from a distribution driven from $W$ and the set of independent parameters,~$\theta$.

For a given layer $l$ of the neural network, the perturbation-injection modules can be used to achieve the following probability model for the layer's final activations:
\begin{align}
    P_{l}(X_l;W_l,\theta_l) = f_l(X_l,W_l) + Q(\theta_l)
    \label{eq:main_eq}
\end{align}
where $f_l(X_l,W_l)$ represents the activation of layer $l$ with weights $W_l$, $X_l$ as its input, and $Q(\theta_l)$ is a noise distribution with parameters $\theta_l$ following an exponential distribution function. While $Q(\cdot)$ can be any exponential distribution, we choose Gaussian distribution because of its simplicity to model and effectiveness, which can be formulated as follow:
\vspace{-0.5cm}
\begin{align}
    Q(\theta_l) = \theta_l \cdot \mathcal{N}(0, 1). 
    \label{eq:quas}
\end{align}
The parameter $\theta_l$ scales the magnitude of the output from the normal distribution encoding the standard deviation of the distribution $Q(\cdot)$.
Substituting the right hand-side of $Q(\cdot)$ defined in \eqref{eq:quas} into \eqref{eq:main_eq} enforces $P_l(\cdot)$ to follow a Gaussian distribution:
\begin{align}
    P_l(X_l;W_l,\theta_l) \approx \mathcal{N} \Big(f_l(X_l,W_l),\theta_l\Big).
    \label{eq:dist_model}
\end{align}
This new probabilistic formulation of layer activations can be extended to the whole network, so instead of a deterministic output $Y$, network outputs \mbox{$P(X;W,\theta) \approx \mathcal{N} \Big(f(X,W),\theta\Big)$}, with $W$ and $\theta$ showing the parameters of all layers. 

 Having this new formulation for a deep neural network, a proper training process to effectively learn both sets of these parameters is highly desired. To this end, we propose a new training mechanism to learn both network parameters and perturbation-injection modules in an  alternating back-propagation approach. 
 
\subsection{Alternating Back-Propagation}
The proposed neural network structure comprises of two sets of parameters, $W$ and $\theta$, being trained given training samples $(X, T)$ as the input and the ground truth output to the network. However, these two sets of parameters are  in conflict with each other and try to push the learning process in two opposite directions. Having the probabilistic representation $P(\cdot)$, $W$ is mapping the input $X$ to output $T$ based on  the mean of the distribution $P(\cdot)$, $f(X, W)$; while, the set of $\theta$ improves the generalization of the model by including perturbations into the training mechanism.

The proposed alternating back-propagation framework decouples the learning process associated to network parameters $W$ and perturbation-injection distributions $\theta$  to effectively update both sets of parameters.  
 To this end, the network parameters and perturbation-injection modules are updated in a consecutive manner.

The training process of the proposed Learn2Perturb is done within two main steps:
\begin{itemize}
    \item \textit{Perturbation-injected network training}; the parameters of the network, $W$, are updated via gradient descent to decrease the network loss in the presence of perturbations, caused by the currently fixed perturbation-injection distribution, $Q|_{\theta}$.
    \item \textit{Perturbation-injection distribution training}; the parameters of the perturbation-injection distribution, $\theta$, are updated  given the set of parameters $W$ are fixed to improve the generalization of the network and as a result, improve its robustness against adversarial perturbation.
\end{itemize}
These two steps are performed consecutively; however, the number of iterations for each step before moving to the next step can be determined based on the application.

\begin{algorithm}[!h]
 \label{alg:training}
 \SetKwInOut{Data}{Input}
  \SetKwInOut{Result}{Output}
  \SetKwInOut{init}{Initialization}
\Data{Training set D = $\{(x_i,t_i),\;\; i = 1, \ldots, n\}$ \\ Number of training epochs, $I$ \\ $\theta_{min}$, the lower bound for $\theta$ \\$\theta_0$, initial values for $\theta$\\Learning rate, lr, and constant $\gamma$}
  \Result{Learned parameters $W$ \\ Learned noise distributions $Q(\theta)$} 
  
 
 \For{ $t \gets 1$ to $I$}{
  \textbf{Perturbation-injected training:} update $W$ based on the loss function $\mathcal{L}(\cdot)$ Eq.~\eqref{eq:loss} while $\theta$ is fixed\\ 
  $W^t \gets W^{t-1} - lr \cdot \nabla_W \mathcal{L}\Big(P(X; W^{t-1}, \theta^{t-1}),T \Big)$ \\
  \textbf{Perturbation-injection module training:}\\ 
  update $\theta$ based on Eq.~\eqref{eq:loss} while $W$ is fixed\\$\theta^t \gets \theta^{t-1} - lr \cdot \nabla_{\theta} \mathcal{L}\Big(P(X; W^{t-1}, \theta^{t-1}),T\Big) - \gamma \cdot \nabla_{\theta} g(\theta^{t-1})$ \\
  Values of $\theta^t$ smaller than $\theta_{min}$ are projected to $\theta_{min}$\\
 }
 
 \caption{Alternating back-propagation of the Learn2Perturb framework}
 
 \end{algorithm}

Utilizing a generic loss function in the training of the network when the perturbation-injection modules are embedded forces the noise parameters to converge to zero and eventually removes the effect of the perturbation-injection distributions by making them very small. In other words, the neural network with generic loss tends to learn $P(\cdot)$ as a Dirac distribution where the $Q(\cdot)$ is close to zero; to prevent the aforementioned problem, a new regularization term is designed and added to  the loss  function.
As such the new loss function can be formulated as:
\begin{align}
\label{eq:loss}
\underset{W,\theta}{\text{arg min}}\Big[ {\mathcal{L}\Big(P(X; W, \theta),T\Big)} + \gamma \cdot {g({\mathcal{\theta})}}\Big]
\end{align}
where $ \mathcal{L}(\cdot)$ is the classification loss function (i.e., usually cross entropy) such that the set of parameters $W$ need to be tuned to generate the associated output of the input $X$.
The function $g(\theta)$ is the regularizer enforcing smooth increase in the parameters $\theta = \{\theta_{l,j}\}^{l=1:K}_{j = 1:M_l}$, where $\theta_{l,j}$ shows the $j$th noise parameter in the $l$th layer, corresponding to an element of the output feature map. $K$ and $M_l$ represent the number of layers and noise parameters per layer, respectively. $\gamma$ is the hyper-parameter balancing the two terms in the optimization. Independent distributions are learnt for perturbation-injection models in each layer. 
 The regularizer function should be enforced with an annealing characteristic where the perturbation-injection distributions are gradually improved and converged  thus the parameters $W$ can be trained effectively.  As such the regularization function is formulated as below:
\begin{align}
    g(\theta) = - {\frac{ \theta^{1/2}}{\tau}} 
    \label{eq:reg} 
\end{align}
where $\tau$ is the output of a harmonic series given the current epoch value in the training process. Using a harmonic series to determine $\tau$, gradually decreases the effect of the reqularizer function in the loss and lets the neural network converge. While the squared root of $\theta$ makes the equation easier to take the derivative, it also provides a slower rate of change for larger values of $\theta$ which helps the network to converge to a steady state smoothly.  

As seen in Algorithm~\ref{alg:training} at first, the perturbation-injection distributions $Q$ and network parameters $W$ are initialized. Then the model parameters $W$ are updated based on the classification loss $\mathcal{L}(\cdot)$, and this loss function is minimized in the presence of perturbation-injection modules. Then, the perturbation-injection distributions $Q$ are updated by performing the ``perturbation-injection module training" step.

One of the main advantages of this approach is that since the learning process of these two sets of parameters is decoupled, the training process can be easily performed without a large amount of manual hyper-parameter tweaking compared to other randomized state-of-the-art approaches. Moreover, the proposed method can help the model to converge faster as the  perturbation-injection distributions are continuously improved during the training process. 


\subsection{Model Setup, Training and Inference}
 Perturbation-injection distributions  are added to the network  in different locations and specifically after each convolution operation to create a new network model based on the Learn2Perturb framework. As shown in Figure~\ref{fig:learn2perturb}, these modules generate the perturbations with the same size as the feature activation maps of that specific layer. Each perturbation-injection distribution follows independent distribution and therefore, the generated perturbation value for each feature is drawn independently.

In the training phase, the model parameters and the perturbation-injection distributions are trained in an iterative and consecutive manner and based on the proposed alternating back-propagation approach. It is worth to mention that the model parameters are trained for 20 epochs before activating the perturbation distributions to help the network parameters converge to a good initial point. After 20 epochs, the alternating back-propagation is applied to train both model parameters and perturbation-injection distributions. Furthermore, we take advantage of adversarial training technique which adds on-the-fly adversarial examples into the training data, to improve the  model's robustness more effectively against perturbations. As such PGD adversarial technique is incorporated in the training to  provide stronger guarantee bounds against all first-order adversaries optimizing in $l_{\infty}$ space. 

The perturbation-injection distributions are applied in the inference step, as well. This will introduce a dynamic nature into the inference process and as a result, it makes it harder for the adversaries to find an optimal adversarial examples to fool the network.

\vspace{-0.25cm}
\section{Experiments}
\label{sec:res}
\vspace{-0.25cm}
To illustrate the effectiveness of the proposed Learn2Perturb, we train various models using this framework and evaluate their robustness against different adversarial attack algorithms. Furthermore, the  proposed method is compared with  different state-of-the-art approaches including PGD adversarial training~\cite{madry2017towards} (also denoted as Vanilla model), Parametric Noise Injection (PNI)~\cite{he2019parametric}, Adversarial Bayesian Neural Network (Adv-BNN)~\cite{liu2018adv}, Random Self-Ensemble (RSE)~\cite{liu2018towards} and PixelDP (DP)~\cite{lecuyer2018certified}.

\vspace{-0.05cm}
\subsection{Dataset \& Adversarial Attacks }
\vspace{-0.05cm}
For the evaluation purpose, the CIFAR-10 and CIFAR-100 datasets\footnote{Experimental results for CIFAR-100 dataset are reported in the supplementary material.}~\cite{krizhevsky2009learning} are utilized for training and evaluating the networks. Both of these datasets contain 50,000 training data and 10,000 test data of natural color images of \mbox{$32\times32$}; however, CIFAR-10 has 10 different class with 6000 images per class, while CIFAR-100 has 100 classes with 600 images per class.  

Different white-box and black-box attacks are utilized to evaluate the proposed Learn2Perturb along with state-of-the-art methods. The competing algorithms are evaluated via white-box attacks including FGSM~\cite{szegedy2013intriguing}, PGD~\cite{kurakin2016adversarial_scale} and C\&W attacks~\cite{carlini2017towards}.  One-Pixel attack~\cite{su2019one}, and Transferability attack~\cite{papernot2016transferability} are utilized as the black-box attacks to evaluate the competing method.

\vspace{-0.15cm}
\subsection{Experimental Setup}
\vspace{-0.05cm}
We use ResNet based architectures~\cite{he2016deep} as the baseline for our experiments; The classical ResNet architecture (i.e., ResNet-V1 and its variations) and the new ResNet architecture (i.e., ResNet-V2) are used for evaluation. The main difference between two architectures is the number of stages and the number of blocks in each stage. Moreover, average pooling  is utilized for down-sampling in ResNet-V1 architecture while the ResNet-V2 uses $1\times1$ CNN layers for this purpose. Followed by the experimental setup proposed in~\cite{he2019parametric}, data normalization is done via adding a non-trainable layer at the beginning of the network and the adversarial perturbations are directly added to the original input data, before normalization being applied.
Both adversarial training and robustness testing setup follow the same configurations as introduced in~\cite{madry2017towards} and~\cite{he2019parametric}. Adversarial training with PGD and testing robustness against PGD, are both done in 7 iterations with the maximum $l_{\infty} = 8/255$ (i.e., $\epsilon$) and step sizes of 0.01 for each iteration. FGSM attack also uses the same 8/255 limit for perturbation. 
For C\&W attack, we use ADAM~\cite{kingma2014adam} optimizer with learning rate $5e^{-4}$. Maximum number of iterations is 1000, and for the constant $c$ in \ref{eq:cw} we choose the range $1e^{-3}$ to $1e^{10}$; furthermore to find the  value of $c$, binary search with up to 9 steps is performed. The confidence, $\kappa$, parameter of C\&W attack, which turns out to have a big effect while evaluating defense approaches involving randomization, takes values ranging from 0 to 5.

In the case of transferability attacks, a PGD adversarially trained network (i.e. a vanilla model) is used as the source network for generating adversarial examples and these adversarial samples are then utilized to attack competing models. For one/few-pixel attacks, we consider the case \mbox{\{1, 2, 3\}-pixel} attack in this work.
\footnote{A more detailed experimental setup is provided in the supplementary material.}

\vspace{-0.25cm}
\begin{table*}
\vspace{-0.15cm}
\caption{Evaluating the effectiveness of the proposed perturbation-injection modules by comparing against adversarial training algorithm (Vanilla) within the proposed framework and its variation (Learn2Perturb-R). }
\vspace{-0.05cm}
\centering
\footnotesize
\setlength{\tabcolsep}{0.075cm} 
\begin{tabular}{lrcccccccccccc}
 \hline
   ~& ~& \multicolumn{3}{c}{\bf No defense} & \multicolumn{3}{c}{\bf Vanilla\cite{madry2017towards}} & \multicolumn{3}{c}{\bf Learn2Perturb-R} & \multicolumn{3}{c}{\bf  Learn2Perturb} \\
   \cmidrule(lr){3-5} \cmidrule(lr){6-8} \cmidrule(lr){9-11} \cmidrule(lr){12-14} 
       \bf Model &\bf  \#Parameter  &
       \bf Clean &\bf  PGD &\bf  FGSM &
       \bf Clean &\bf  PGD & \bf FGSM &
       \bf Clean &\bf  PGD &\bf  FGSM &
       \bf Clean &\bf  PGD &\bf  FGSM \\
          \hline
       
       \bf ResNet-V1(20)   &   269,722 &92.1 &  0.0$\pm$0.0 &  14.1 &   83.8 &  39.1$\pm$0.1 &  46.6 & 81.15$\pm$0.02 & 50.23$\pm$0.14 & 55.89$\pm$0.04 & 83.62$\pm$0.02 & 51.13$\pm$0.08 & 58.41$\pm$0.07
       \\  

     \bf ResNet-V1(56)   &   853,018 &  93.3 &  0.0$\pm$0.0 &  24.2  &  86.5 &  40.1$\pm$0.1  & 48.8 & 82.35$\pm$0.03 &  53.30$\pm$0.10 & 58.71$\pm$0.04 & 84.82$\pm$0.04 & 54.84$\pm$0.10  & 61.53$\pm$0.04
       \\
       \bf ResNet-V2(18) & 11,173,962 &  95.2 & 0.1$\pm$0.0  & 43.1  &   85.46 & 43.9$\pm$0.0 &  52.5 &  82.46$\pm$0.17&  53.33$\pm$0.12   & 59.09$\pm$0.17 & 85.30$\pm$0.09 & 56.06$\pm$0.16& 62.43$\pm$0.06
       \label{tab:learn2perturb_exp}
\end{tabular}

\end{table*}

\vspace{-0.05cm}
\subsection{Experimental Results}
\vspace{-0.2cm}
To evaluate the proposed Learn2Perturb framework, the  method is compared with PGD adversarial trained model (also denoted as Vanilla). The proposed module is evaluated on three different ResNet architectures. Table~\ref{tab:learn2perturb_exp} shows the effectiveness of the proposed Learn2Perturb method in improving the robustness of different networks architectures.  Results demonstrate that the proposed  perturbation-injection modules improve the network's robustness. As seen, the proposed perturbation-injection modules can provide robust performance on both `ResNet-V1' (both with 20 and 56 layers) and `ResNet-V2' (18 layers) architectures against both FGSM and PGD attacks which illustrates the effectiveness of the proposed  module in providing more robust network architectures. 
Furthermore, the evaluation results for no defense approach (a network without any improvement) are provided as a reference point.

We also evaluate a variation of the proposed Learn2Perturb framework (i.e. Learn2Perturb-R) where we analyze a different approach in performing the two steps of ``perturbation-injected network training" and ``perturbation-injection module training". In this variation, the perturbation-injection modules are only updated using the regularizer function $g(\theta)$, and network gradients are not used to update $\theta$ parameters. 

As it can be seen in table~\ref{tab:learn2perturb_exp}, taking advantage of both network gradient and the regularizer performs better than when we only take into account the  regulizer effect. One reason to justify this outcome is allowing the gradient of loss function $\mathcal{L}(\cdot)$ to update perturbation-injection modules in Learn2Perturb. This would let the loss function  to react to perturbations when they cannot tolerate the injected noise and updates the perturbation-injection noise modules more frequently. Nonetheless, the results in table~\ref{tab:learn2perturb_exp} show that Learn2Perturb-R still outperforms other proposed methods in adversarial robustness, though it suffers from smaller clean data accuracy.

\begin{figure}
\vspace{-0.8cm}
    \centering
    \includegraphics[width=6cm]{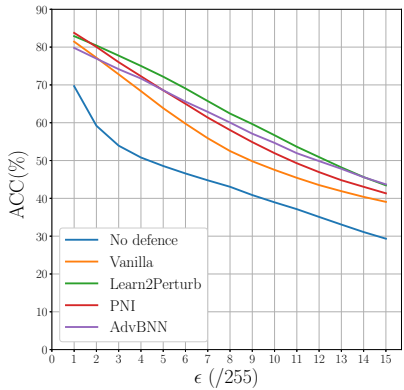}
    \vspace{-0.25cm}
    \caption{Analyzing the effectiveness of the proposed method compared to state-of-the-art algorithms on different $\epsilon$ values for FGSM attack.}
    \vspace{-0.15cm}
    \label{fig:fgsm_variation}
\end{figure}

\subsection{ Robustness Comparison}
In this section, to further illustrate the effectiveness of the proposed Learn2Perturb framework, we compare Learn2Perturb with PNI~\cite{he2019parametric} and Adv-BNN~\cite{liu2018adv} as two randomization state-of-the-art  approaches to improve the robustness of deep neural networks. Table~\ref{tab:net_cap} reports these comparison results for different network architectures varying in network depth and capacity. We examine the effect of different network depths including ResNet-V1(20), ResNet-V1(32), ResNet-V1(44) and ResNet-V1(56)  along with the effect of network width in Table~\ref{tab:net_cap}  by increasing the number of filters in ResNet-V1(20) which results to  ResNet-V1(20)[1.5$\times$], ResNet-V1(20)[2$\times$] and ResNet-V1(20)[4$\times$]. As seen, while the competing methods do not provide consistent performance by increasing the capacity of the network (increasing depth or width) the proposed framework provides consistent robustness through different network capacities.

\begin{table*}
\vspace{-0.25cm}
\caption{The effect of network capacity on the performance of the proposed method and other state-of-the-art algorithms. The proposed Learn2Perturb is compared  with Parametric Noise Injection (PNI) method~\cite{he2019parametric}  and Adv-BNN~\cite{liu2018adv}. Results shows the effectiveness of the  proposed Learn2Perturb algorithm in training robust neural network models. To have a fair comparison, we evaluated methods on different network sizes and capacities.  Result are reported by standard deviation because of the randomness involved in these methods. }
\centering
\footnotesize
\setlength{\tabcolsep}{0.1cm} 
\begin{tabular}{lrccccccccc}
 \hline
   ~& ~&  \multicolumn{3}{c}{\bf PNI~\cite{he2019parametric}} &
   \multicolumn{3}{c}{\bf Adv-BNN~\cite{liu2018adv}} & \multicolumn{3}{c}{\bf  Learn2Perturb} \\
   \cmidrule(lr){3-5} \cmidrule(lr){6-8}
   \cmidrule(lr){9-11}
       \bf Model &\bf  \#Parameter  &
       \bf Clean &\bf  PGD &\bf  FGSM &
       \bf Clean &\bf  PGD &\bf  FGSM &
       \bf Clean &\bf  PGD & \bf FGSM \\
    
          \hline
       \bf ResNet-V1(20)   &   269,722  &    84.90$\pm$0.1 & 45.90$\pm$0.1 & 54.50$\pm$0.4 & 65.76$\pm$5.92 & 44.95$\pm$1.21 & 51.58$\pm$1.49 & 
       83.62$\pm$0.02 & 51.13$\pm$0.08 & 58.41$\pm$0.07
       \\  
       
       \bf ResNet-V1(32)   &   464,154  &     85.90$\pm$0.1 & 43.50$\pm$0.3 & 51.50$\pm$0.1 &
       62.95$\pm$5.63 & 54.62$\pm$0.06 & 50.29$\pm$2.70 &
       84.19$\pm$0.06 & 54.62$\pm$0.06 & 59.94$\pm$0.11
       \\ 
       \bf ResNet-V1(44)   &   658,586  &    84.70$\pm$0.2 & 48.50$\pm$0.2 & 55.80$\pm$0.1 & 76.87$\pm$0.24 & 54.62$\pm$0.06 & 58.55$\pm$0.49 &
       85.61$\pm$0.01 & 54.62$\pm$0.06 & 61.32$\pm$0.13
       \\ 
       
       \bf ResNet-V1(56)   &   853,018  &    86.80$\pm$0.2 & 46.30$\pm$0.3 & 53.90$\pm$0.1 & 77.20$\pm$0.02 & 54.62$\pm$0.06 & 57.88$\pm$0.02 &
       84.82$\pm$0.04 & 54.62$\pm$0.06 & 61.53$\pm$0.04
       \\ \hline
       
       \bf ResNet-V1(20)[1.5$\times$] &  605,026  &    86.00$\pm$0.1 & 46.70$\pm$0.2 & 54.50$\pm$0.2 & 65.58$\pm$0.42 & 28.07$\pm$1.11 & 36.11$\pm$1.29
       & 85.40$\pm$0.08 & 53.32$\pm$0.02 & 61.10$\pm$0.06
       \\
       \bf ResNet-V1(20)[2$\times$] &  1,073,962 &     86.20$\pm$0.1 & 46.10$\pm$0.2 & 54.60$\pm$0.2 & 79.03$\pm$0.04 & 53.46$\pm$0.06 & 58.30$\pm$0.14 & 85.89$\pm$0.10 & 54.29$\pm$0.02 & 61.61$\pm$0.05
       \\
      
       \bf ResNet-V1(20)[4$\times$] &  4,286,026  &   87.70$\pm$0.1 & 49.10$\pm$0.3 & 57.00$\pm$0.2& 82.31$\pm$0.03 & 52.61$\pm$0.12 & 59.01$\pm$0.04
       & 86.09$\pm$0.05 & 55.75$\pm$0.07 & 61.32$\pm$0.02
       \\ \hline
       
       \bf ResNet-V2(18) & 11,173,962 &   87.21$\pm$0.00  &  49.42$\pm$0.01 & 58.06$\pm$0.02
       & 82.15$\pm$0.06 & 53.62$\pm$0.06& 60.04$\pm$0.01& 85.30$\pm$0.09 & 56.06$\pm$0.08& 62.43$\pm$0.06
       \label{tab:net_cap}
\end{tabular}
\end{table*}


\begin{table}
\vspace{-0.5cm}
\caption{Comparison results of the proposed Learn2Perturb and competing methods on C$\&$W~\cite{carlini2017towards} attack.}
\centering
\footnotesize
\setlength{\tabcolsep}{0.1cm} 
\begin{tabular}{l|ccccc}
       \bf Confidence &\bf  No defense & \bf Vanilla &\bf  PNI &\bf Adv-BNN &\bf  Learn2Perturb\\
          \hline
       
       \bf $\kappa=0.0$ &  0.0 & 0.0 & 66.9 & 78.9 &\bf 83.6\\  
       
       \bf $\kappa=0.1$   &  0.0 & 0.0 & 66.1 &78.1 &\bf 84.0\\ 
       
       \bf $\kappa=1.0$   &  0.0 & 0.0& 34.0 &65.1 &\bf 76.4\\
       
       \bf $\kappa=2.0$   &  0.0 & 0.0& 16.0 & 49.1&\bf 66.5\\
       
       \bf $\kappa=5.0$   &  0.0 & 0.0& 0.8 &16.0 &\bf 34.8\\
       
       \label{tab:cw_attack}
\end{tabular}
\end{table}

\begin{table}
\vspace{-0.45cm}
\caption{Comparison results of the proposed Learn2Perturb and state-of-the-art methods in providing a robust network model. Some of the numbers are extracted from~\cite{he2019parametric}. The reported results are either based on the maximum accuracy achieved  in the literature or own results if we achieved higher level of accuracy.}
\centering
\footnotesize
\setlength{\tabcolsep}{0.1cm} 
\begin{tabular}{lccc}

       \bf Defense Method &\bf  Model & \bf Clean &\bf  PGD \\
          \hline
       
       \bf RSE~\cite{liu2018towards} &  ResNext & 87.5 & 40 \\  
       
       \bf DP~\cite{lecuyer2018certified}   &  28-10 wide ResNet & 87 & 25 \\ 
       
       \bf Adv-BNN~\cite{liu2018adv}   & ResNet-V1(56) & 77.20& 54.62$\pm$0.06 \\ 
       
       \bf PGD adv. training~\cite{madry2017towards} &  ResNet-V1(20) [4$\times$] & 87 & 46.1$\pm$0.1 \\ 
       
       \bf PNI~\cite{he2019parametric}   &  ResNet-V1(20) [4$\times$] &\bf  87.7$\pm$0.1& 49.1$\pm$0.3 \\ 
       \bf Learn2Perturb &  ResNet-V2(18)  & 85.3$\pm$0.1& \bf 56.3$\pm$0.1 \\ 

       \label{tab:full_compare}
\end{tabular}
\vspace{-0.7cm}
\end{table}


       
       
       
       
       


\begin{figure}
\vspace{-0.5cm}
    \centering
    \includegraphics[width=6cm]{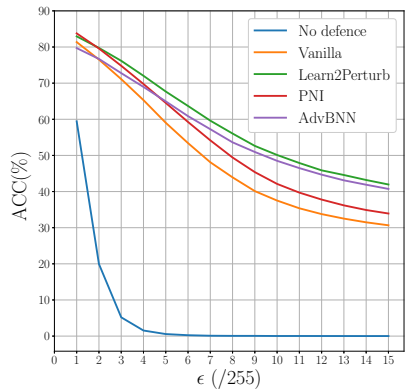}
    \vspace{-0.25cm}
    \caption{Evaluating the robustness of the proposed Learn2Perturb compared with other state-of-the-art methods through different $\epsilon$ based on PGD attack.}
    \vspace{-0.35cm}
    \label{fig:pgd_variation}
\end{figure}


The reported results  in Table~\ref{tab:net_cap} show that while PNI provides minor  boosting in network accuracy on  clean data, the proposed Learn2Perturb method performs with much higher accuracy when the input data is perturbed with adversarial noise. The main reason for this phenomena is the fact that PNI reach to a very low level of the noise perturbation during the training as the loss function tries to remove the effect of perturbation by making the noise parameters to zero. The results  demonstrate that the proposed Learn2Perturb algorithm outperforms the PNI method by 4-7\% on both FGSM and PGD adversarial attacks. 
The proposed method is also compared with Adv-BNN~\cite{liu2018adv}. Results show that while Adv-BNN can provide robust networks in some cases compared to PNI, it is not scalable when the network width is increased and the performance of the networks drop drastically. This is illustrated one of the drawbacks of Bayesian approach which they need to be designed carefully for each network architecture separately.

It has been shown, there is no guarantee that methods robust against $l_\infty$ attacks would provide same level of robustness against $l_2$ attacks~\cite{araujo2019robust}. Araujo {\it et al.}~\cite{araujo2019robust} illustrated experimentally that randomization technique trained with $l_{\infty}$ can improve the robustness against $l_2$ attacks as well. In this work we further validate this finding. In order to provide more powerful $l_2$ attacks  challenging the effect of randomization, we apply C\&W attacks with different confidence values, $\kappa$. The parameter $\kappa$ enforces the $f(\cdot)$ in \ref{eq:cw} to be $\leq-\kappa$ rather than simply $\leq$ 0. As seen in Table~\ref{tab:cw_attack}, for bigger values of $\kappa$ the success rate of C\&W attack increases; nonetheless, our proposed method outperforms the other competing methods with a big margin for all values of~$\kappa$.

Table~\ref{tab:full_compare} shows the comparison results for the proposed method and state-of-the-art approaches in providing robust network model on CIFAR-10 dataset. The proposed Learn2Perturb method outperforms other state-the-art methods and provides a more robust network model with better performance when dealing with PGD attack.

We also analyze the effectiveness of the proposed method in dealing with different adversarial noise levels. To this end, the ResNet-V2(18) architecture is utilized for all competing methods. The network architectures are designed and trained via four different competing methods; and the trained networks are examined with both FGSM and PGD attacks but with a variation of $\epsilon$ values.

Figure~\ref{fig:fgsm_variation} demonstrates the robustness of four competing methods in dealing with FGSM adversarial attack. As seen, while increasing  $\epsilon$ decreases the robustness of all trained  networks, the network designed and trained by the proposed Learn2Perturb approach outperforms other methods through all variations of adversarial noise values ($\epsilon$'s).

To confirm the results shown in Figure~\ref{fig:fgsm_variation}, the same experiment is conducted to examine the robustness of the trained networks on PGD attack. While the PGD attack is more powerful in fooling the networks, results show that the network designed and trained by the proposed Learn2Perturb framework still outperforms other state-of-the-art approaches.

\vspace{-0.09cm}
\subsection{Expectation over Transformation (EOT)}
\vspace{-0.075cm}
Athalye {\it et. al}~\cite{athalye2018obfuscated} showed that many of the defense algorithms that take advantage of injecting randomization to network interior layers or applying random transformations on the input before feeding it to the network achieve robustness through false stochastic gradients. They further stated that these methods obfuscate the gradients that attackers utilize to perform iterative attacking optimizations. As such, they proposed the EOT attack (originally introduced in~\cite{athalye2017synthesizing}) to evaluate these types of defense mechanisms. They showed that the false gradients cannot protect the network when the attack uses the gradients which are the expectation over a series of transformations.  

Since our Learn2Perturb algorithm and other competing methods involve randomization, the tested algorithms in this study are evaluated via the EOT attack method as well. To do so, followed by~\cite{pinot2019theoretical}, at every iteration of PGD attack, the gradient is achieved as the expectation calculated from a Monte Carlo method with 80 simulations of different transformations. Results show that the network trained via PNI can provide 48.65\% robustness compared to Adv-BNN which provides 51.19\% robustness for the CIFAR-10 dataset against this attack. Experimental result illustrates that the proposed Learn2Perturb approach can produce a model which achieves 53.34\% robustness and outperforms the other two state-of-the-art algorithms.

It is worth mentioning that the experimental results showed that neither the proposed Learn2Perturb method nor the other competing approaches studied in this work suffer from obfuscated gradients. Furthermore, the proposed Learn2Perturb method successfully passes the five metrics introduced in~\cite{athalye2018obfuscated}, and thus further illustrates that Learn2Perturb is not subjected to obfuscated gradients.

\vspace{-0.3cm}
\section{Conclusion}
\vspace{-0.2cm}
In this paper, we proposed Learn2Perturb, an end-to-end feature  perturbation learning approach for improving adversarial robustness of deep neural networks. Learned perturbation injection modules are introduced to increase  uncertainty during both training and inference to make it harder to craft successful adversarial attacks. A novel alternating back-propagation approach is also introduced to learn both network parameters and perturbation-injection module parameters in an alternating fashion. Experimental results on both different black-box and white-box attacks demonstrated the efficacy of the proposed Learn2Perturb algorithm, which outperformed the state-of-the-art methods in improving robustness against different adversarial attacks.  Future work involves exploring extending the proposed modules to inject a greater perturbation type diversity for greater generalization in terms of adversarial robustness.


\input{ms.bbl}
\include{Supplementary}

\end{document}

%% file: Supplementary.tex



\cvprfinalcopy 





\appendix
\hspace{-5mm} {\huge Supplementary Material}

\section{Detailed Analysis}
Here we provide a more detailed analysis of the experiments evaluating the proposed method and other tested algorithms with regards to their theoretical background, training, and evaluation.
\subsection{Embedding perturbation-injection modules in  a  network} 
Generally perturbation-injection modules can be embedded after the activation of each layer. However, for the ResNet baselines we choose to add them just to the output of every block and before the ReLU activation. We do this to reduce the amount of trainable parameters and reduce the training and inference times. Nevertheless, any other setup can be used as well.

\subsection{Behaviour of noise distributions in PNI vs Learn2Perturb}
As stated in Section 3.2, the trained noise parameters by the PNI approach fluctuate during the training because of the loss function. The min-max optimization applied in that methodology causes the training to enforce noise parameters to be zero as the number of training epoch increases. As such, it is crucial to select the right number of epochs in the training step.

This issue has been addressed in the proposed Learn2Perturb algorithm by introducing a new regularization term in the loss function. As a result, there is a trade-off between training proper perturbation-injection distribution and modeling accuracy during the training step. This trade-off would let the perturbation modules to learn properly and eventually converge to a steady state.
 To this end, a harmonic series term is introduced in the proposed regularization term  which decreases the effect of regularization as the number of training epochs increases, and help the perturbation-injection modules to converge.
 
\begin{figure}
\vspace{-0.5cm}
    \centering
    \includegraphics[width=7cm]{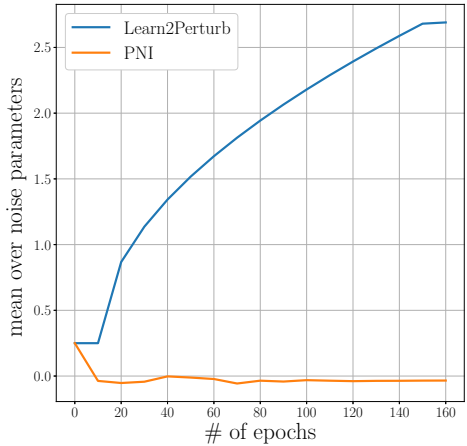}
    \vspace{-0.25cm}
    \caption{Evolution of mean over noise perturbation parameters through training epochs for ResNet V2. As seen, while the noise distributions are growing the in the Learn2Perturb algorithm, they converge close to zero in the PNI method.  }
    \label{fig:track_noise}
\end{figure}

 Figure \ref{fig:track_noise} shows the behaviour of noise distributions in both PNI and Learn2Perturb algorithm during the training. As seen, the proposed Learn2Perturb algorithm can handle the noise distributions properly and as a result,  the noise distribution parameters are being trained as the number of training epoch increases until they converge to a steady state. However, the noise distributions are forced to zero for the model trained via the PNI algorithm due to the way the loss function is formulated.  
 


\subsection{Theoretical Background}
It has been illustrated by Pinot {\it et al.}~\cite{pinot2019theoretical} that randomizing a deep neural network can improve the robustness of the model against adversarial attacks.  
A deep neural network $M$ is a probabilistic mapping when it maps $\mathcal{X}$ to $\mathcal{Y}$ via \mbox{$M: \mathcal{X} \xrightarrow{}P(\mathcal{Y})$}; to obtain a numerical output  of this probabilistic mapping, one needs to sample $y$ according to $M(x)$.

The probabilistic mapping $M(x)$ is $d_{P(Y)}(\alpha, \epsilon, \gamma)$ robust if PC-Risk$_\epsilon(M,\epsilon) \leq \gamma$, where PC-Risk$_\epsilon(M,\epsilon)$ is defined  as the minimum value of $\tau$ when \mbox{$d_{P(Y)}\Big(M(x+t), M(x)\Big) > \epsilon$} and $d_{P(Y)}(\cdot)$ is a metric/divergence on $p(\mathcal{Y})$.
If $M(x)$ follows an Exponential family distribution, it is possible to define the upper bound for the robustness of the model based on $\epsilon$-perturbation.


\subsection{Detailed Experimental Setup}
In order to encourage the reproducible experimental results, in this section we provide a detailed explanation of the experimental setup and environment of the reported experiments.
Pytorch version 1.2 was used for developing all experiments, and our codes will be open sourced upon the acceptance of this paper.

Following the observation made by Madry {\it et al.}~\cite{madry2017towards}, capacity of networks alone can help increasing the robustness of the models against adversarial attacks. As such, we compare Learn2Perturb and competing state-of-the-art methods for various networks with different capacities. 

The ResNet~\cite{he2016deep} architectures has been selected as the baseline network followed by the state-of-the-art methods and the fast convergence property of this network. The effect of network depth were evaluated by examining the competing methods via ResNet-V1(32), (44), (56) as well as ResNet-V1(20) where (x) shows the depth of the network. Moreover, the effect of network width is examined similar to the work done by Zagoruyko and Komodakis~\cite{zagoruyko2016wide}.
To increase the width of the network (i.e, experiment performed on ResNet-V1(20)), the number of input and output channels of each layer is increased by a constant multiplier, $\times 1.5$, $\times 2$, and $\times 4$ which widen the ResNet architecture. However we do not follow the exact approach of \cite{zagoruyko2016wide} in which they applied dropout layers in the network; instead we just increase the width of the basic convolution at each layer by increasing the number of input/output channels. 

We also consider a ResNet-V2(18), which has a very large capacity compared to  ResNet-V1 architecture. Not only the number of channels have increased in this architecture but also it uses $1\times1$ convolutions to perform the down-sampling at each residual blocks.

The proposed Learn2Perturb, No defence, and Vanilla methods,  used the same setup for gradient descent optimizer. SGD optimizer with momentum of 0.9 with Nesterov momentum and weight decay of $1e^{-4}$ is used for training of those methods. The noise injection parameters have weight decay equal to 0. We use the batch size of 128, and 350 epochs to train the model. The initial learning rate is 0.1, then changes to 0.01 and 0.001 at epochs 150 and 250, respectively.

For the parameter $\gamma$ in equation \eqref{eq:loss}, we choose value $10^{-4}$ for all of our experiments. In equation 8, we have $\tau$ which as we state is the output of a harmonic series given the epoch number. we formulate $tau$ as below:
\begin{align}
\label{eq:tau_eq}
\tau(t) = \sum_{i=s}^{t}\frac{1}{i-s-1}
\end{align}
where $t$ shows the current epoch, while $s$ shows the first epoch number from which noise is being added to the network. 

For training models with PNI, the same parameters reported by authors~\cite{he2019parametric} are used. 


The PGD adversarial training utilized alongside with the alternative back-propagation technique in the proposed method which can be formulated as:
\begin{align}
\label{eq:pgd_training}
\underset{W, \theta}{\text{arg min}} \Big[\underset{\delta \in l_{\infty}-\epsilon}{\text{arg max}}  \mathcal{L}\Big(P(X+\delta; W, \theta), T\Big)\Big]
\end{align}
where $W$ encodes the network parameters and $\theta$ shows the perturbation-injection parameters. In this formulation only adversarially generated samples are used in the training step for the outer minimization, following the original work introduced in~\cite{madry2017towards}.

Finally, in order to balance between the adversarial robustness and clean data accuracy~\cite{he2019parametric, zhang2019theoretically}, we formulate the adversarial training as follow:
\begin{align}
\label{eq:pgd_training_balanced}
\underset{W, \theta}{\text{arg min}} \Big[& \alpha \cdot \mathcal{L}\Big(P(X; W, \theta), T\Big) +  \nonumber \\ &\beta \cdot \underset{\delta \in l_{\infty}-\epsilon}{\text{arg max}}  \mathcal{L}\Big(P(X+\delta; W, \theta), T\Big)\Big]
\end{align}
where the first term shows the loss associated to the clean data and $\alpha$ is the weight for the clean data loss term, while the second shows the loss associated with the adversarially generated data with weight $\beta$. The models trained with the proposed Learn2Perturb algorithm use $\alpha = \beta = 0.5$. \eqref{eq:pgd_training_balanced} helps gain adversarial robustness, while maintaining a reasonably high clean data accuracy.

\begin{table*}
\caption{Few-pixel attack; the competing methods are evaluated via few-pixel~\cite{su2019one} attack base on two network architectures of ResNet-V1(20) and ResNet-V2(18). \{1,2,3\} pixels are changed in the test samples to perturbed the images. }
\centering
\footnotesize
\begin{tabular}{lc|ccccc}
       \bf Network Architecture & \bf Attack Strength &\bf  No defence & \bf Vanilla &\bf  PNI &\bf Adv-BNN &\bf  Learn2Perturb\\
          \hline
       \bf \multirow{3}{*}{ResNet-V1(20)} &     \bf 1-pixel &  21.45 & 65.20 & 67.40 & 58.40 & \bf 70.15\\  
       &\bf 2-pixel   &  2.55 & 48.35 & 61.75 & 56.20 & \bf 63.90\\
       &\bf 3-pixel   &  1.10 & 36.40& 58.10 & 55.70 & \bf 61.85\\ 
       \hline
       \bf \multirow{3}{*}{ResNet-V2(18)} &     \bf 1-pixel &  23.44 & 56.10 & 50.90 & \bf68.60 & 64.45\\  
       &\bf 2-pixel   &  3.20 & 33.20 & 39.00 & \bf64.55 & 60.05 \\ 
       &\bf 3-pixel   &  0.95 & 23.95& 35.40 & \bf59.70 & 53.90\\ 
       
       \label{tab:pixel_attack}
\end{tabular}
\end{table*}

\subsection{Black-Box Attacks}
In this section, the robustness of the proposed method and the competing algorithms against black-box attacks are evaluated. Two different attacks including few-pixel attack~\cite{su2019one} and transferability attack~\cite{papernot2016transferability} are used to evaluate the competing methods.

Few-pixel attack (here in the range of one to three pixels) utilizes differential evolution technique to fool deep neural networks under the extreme limitation of only altering at most few pixels. We use population size of 400 and maximum iteration steps of 75 for the differential evolution algorithm. The attack strength is controlled by the number of pixels that are allowed to be modified. In this comparison we consider the \{1,2,3\}-pixel attacks.

\begin{table*}
\caption{Transferability attack comparison. The competing methods are attacked within the context of Transferability where the perturbed images utilized to evaluate the robustness of the model are generated  by one another method. The `Source Model' is the model which the perturbed samples are generated from to attack each competing algorithm. }
\centering
\footnotesize
\setlength{\tabcolsep}{0.15cm} 
\begin{tabular}{ll|cccccccc}
        \hline
       ~& ~& \multicolumn{2}{c}{\bf Vanilla}  & \multicolumn{2}{c}{\bf PNI} & \multicolumn{2}{c}{\bf Adv-BNN} & \multicolumn{2}{c}{\bf Learn2Perturb} \\
       
       \cmidrule(lr){3-4} \cmidrule(lr){5-6} \cmidrule(lr){7-8} \cmidrule(lr){9-10} 
       \bf Network Architecture & \bf Source Model & \bf FGSM & \bf PGD & \bf FGSM & \bf PGD & \bf FGSM & \bf PGD & \bf FGSM & \bf PGD \\
          \hline
       
       \bf \multirow{4}{*}{Resnet20 - V1} & \bf Vanilla &  -- & -- & 60.32$\pm$0.05 & 58.27$\pm$0.01 & 49.22$\pm$0.90 & 48.63$\pm$3.10 & 58.86$\pm$0.03 & 56.75 $\pm$ 0.01 \\  
       
       & \bf PNI &  66.31$\pm$0.02 & 63.04$\pm$0.01 & -- & -- & 51.12$\pm$1.22 & 49.59$\pm$0.83 & 63.26$\pm$0.10 & 59.31 $\pm$ 0.06 \\ 
       
       & \bf Adv-BNN & 74.38$\pm$0.16 & 72.02$\pm$0.02 & 73.05$\pm$0.12 & 70.26$\pm$0.05 & -- & -- & 72.48$\pm$0.05 & 69.25 $\pm$ 0.06 \\
       
       & \bf Learn2Perturb &  70.66$\pm$0.01 & 67.32$\pm$0.01 & 68.46$\pm$0.03 & 64.77$\pm$0.01 & 54.16$\pm$2.36 & 52.23$\pm$1.52 & -- & -- \\
       \hline
       
       \bf \multirow{4}{*}{Resnet18 - V2} & \bf Vanilla &  -- & -- & 69.52$\pm$0.02 & 68.01$\pm$0.02 & 67.20$\pm$0.04 & 65.88$\pm$0.03 & 67.32$\pm$0.04 & 65.58 $\pm$ 0.04 \\  
       
       & \bf PNI &  69.56$\pm$0.01 & 67.09$\pm$0.02 & -- & -- & 67.63$\pm$0.03 & 64.87$\pm$0.04 & 67.63$\pm$0.05 & 64.61 $\pm$ 0.01 \\ 
       
       & \bf Adv-BNN & 73.66$\pm$0.01 & 71.23$\pm$0.01 & 74.02$\pm$0.03 & 71.51$\pm$0.02 & -- & -- & 71.67$\pm$0.09 & 68.75 $\pm$ 0.04 \\
       
       & \bf Learn2Perturb &  73.49$\pm$0.01 & 70.44$\pm$0.00 & 73.89$\pm$0.04 & 70.61$\pm$0.01 & 70.22$\pm$0.04 & 67.33$\pm$0.04 & -- & -- \\
       \label{tab:transferability}
\end{tabular}
\end{table*}

Table~\ref{tab:pixel_attack} shows the comparison results of the competing methods against few-pixel attack.
Two different network architectures (ResNet-V1(20) and ResNet-v2(18)) are used to evaluate the competing algorithms. 
As seen, the proposed Learn2Perturb method outperforms other state-of-the-art methods when the baseline network architecture is ResNet-V1(20). However, Adv-BNN provides better performance when the baseline network architectures is ResNet-V2(18), while the proposed Learn2Perturb algorithm provides comparable performance for this baseline.  


Table~\ref{tab:transferability} demonstrates the comparison results for the proposed Learn2Perturb and state-of-the-art methods  based on Transferability attack. Results again show that the proposed Learn2Perturb method provides robust prediction against this attack as well.

\subsection{CIFAR-100}
\begin{figure}
    \centering
    \includegraphics[width=.5\textwidth]{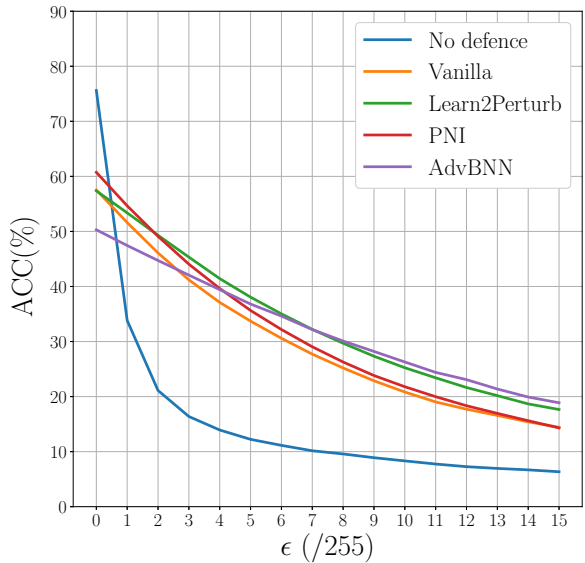}
    \caption{FGSM attack on CIFAR-100 with different epsilons for the $l_{\infty}$ ball on ResNet-V2(18).}
    \label{fig:cifar100-FGSM}
\end{figure}

\begin{figure}
    \centering
    \includegraphics[width=.5\textwidth]{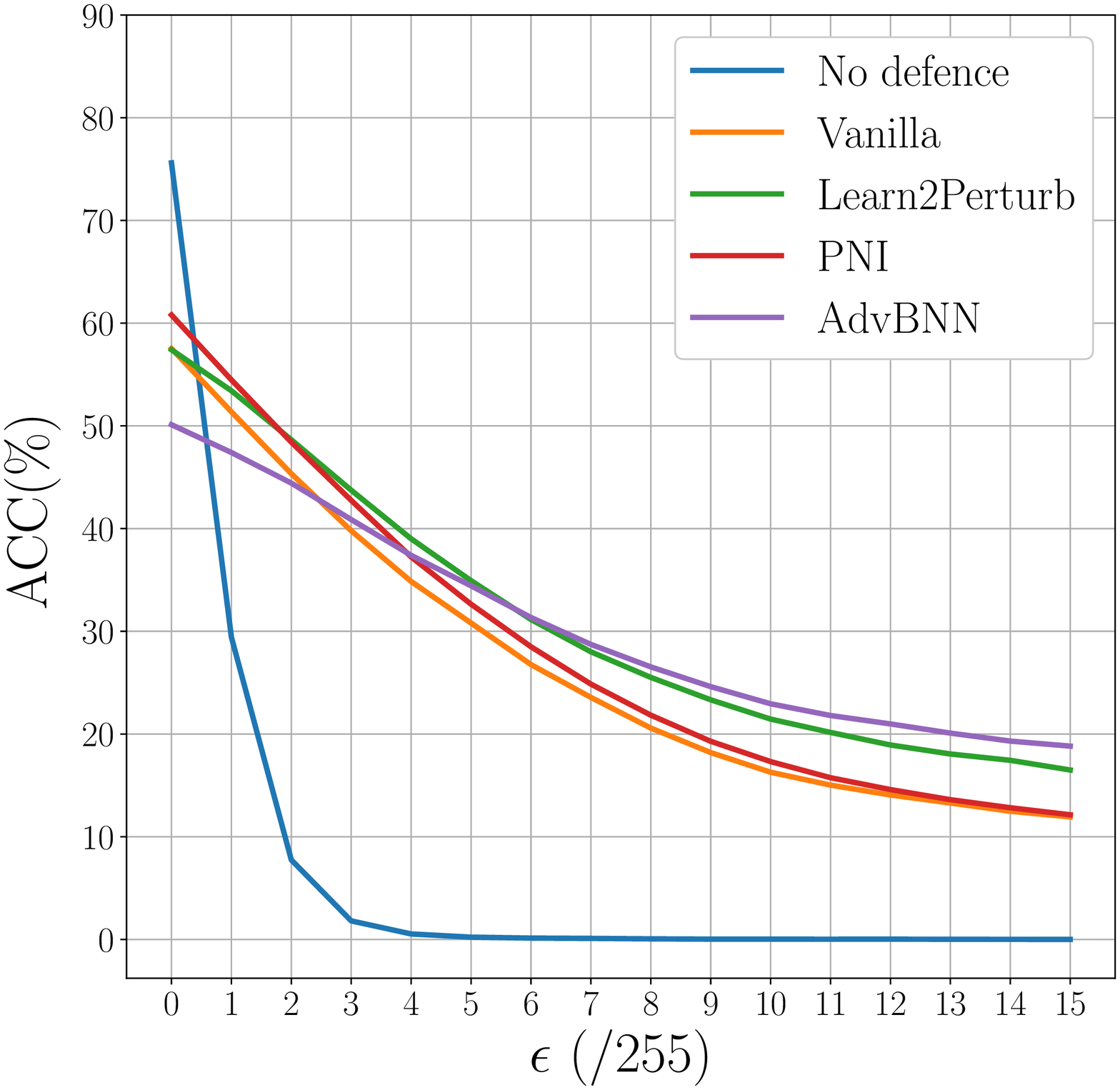}
    \caption{PGD attack on CIFAR-100 with different epsilons for the $l_{\infty}$ ball on ResNet-V2(18).}
    \label{fig:cifar100-PGD}
\end{figure}

A more detailed analysis of the experimental setup and results for the CIFAR-100 dataset is provided as follows.  
The CIFAR-100 dataset~\cite{krizhevsky2009learning} is very similar to CIFAR-10 dataset, however the image samples are categorized to  100 fine class labels. 
All the models involving PGD adversarial training are trained with $\epsilon=\frac{8}{255}$ during training.
Figures~\ref{fig:cifar100-FGSM} and~\ref{fig:cifar100-PGD}  demonstrate the performance comparison of the proposed Learn2Perturb with other state-of-the-art methods on CIFAR-100 dataset based on FGSM and PGD attacks.

As seen, the proposed Learn2Perturb method outperforms other competing algorithms for $\epsilon$s up to $\frac{8}{255}$, however for bigger $\epsilon$s it provides comparable performance with Adv-BNN, which has the best result.

